\pdfoutput=1
\documentclass[11pt]{article}

\usepackage[preprint]{acl}

\usepackage{times}
\usepackage{latexsym}

\usepackage[T1]{fontenc}
\usepackage[utf8]{inputenc}

\usepackage{microtype}

\usepackage{inconsolata}

\usepackage{graphicx}
\usepackage{geometry}
\usepackage{multirow}
\usepackage{multicol}
\usepackage{amsmath} 
\usepackage{algorithm}
\usepackage{algorithmic}

\title{Entropy-Based Block Pruning for Efficient Large Language Models}
\author{Liangwei Yang, Yuhui Xu\thanks{Corresponding author: \texttt{yuhui.xu@salesforce.com}}, Juntao Tan, Doyen Sahoo, Silvio Savarese \\
\textbf{Caiming Xiong, Huan Wang, Shelby Heinecke} \\
Salesforce AI Research, USA
}

\begin{document}
\maketitle
\begin{abstract}
As large language models continue to scale, their growing computational and storage demands pose significant challenges for real-world deployment. In this work, we investigate redundancy within Transformer-based models and propose an entropy-based pruning strategy to enhance efficiency while maintaining performance. Empirical analysis reveals that the entropy of hidden representations decreases in the early blocks but progressively increases across most subsequent blocks. This trend suggests that entropy serves as a more effective measure of information richness within computation blocks. Unlike cosine similarity, which primarily captures geometric relationships, entropy directly quantifies uncertainty and information content, making it a more reliable criterion for pruning. Extensive experiments demonstrate that our entropy-based pruning approach surpasses cosine similarity-based methods in reducing model size while preserving accuracy, offering a promising direction for efficient model deployment. 
\end{abstract}

\section{Introduction}

The emergence of large language models (LLMs) has reshaped current research landscape as well as empowering applications~\citep{dubey2024llama,achiam2023gpt,team2024gemini}. Scaling in size, they demonstrate remarkable performance across a wide range of domains/tasks such as chatbot~\cite{achiam2023gpt}, code generation~\cite{nijkamp2022codegen}, recommendation~\cite{liang2024taxonomy,zhang2025llminit}, etc. Hidden behind these striking achievement, Transformer-based models~\cite{waswani2017attention,touvron2023llama,jiang2023mistral,xue2024xgen} scale their parameter size from millions to billions and research continues to explore even larger architectures~\cite{liu2024deepseek} to further enhance their capabilities. However, the increasing scale in sizes result in substantial computational and storage costs, posing significant challenges for real world deployment.

Recent researches have detected the inherent redundancy of these pre-trained LLMs, especially on the layer level~\cite{gromov2024unreasonable,men2024shortgpt,yang-etal-2024-laco}. Models can maintain competitive performance even after a significant number of layers are removed, indicating that not all layers contribute equally to the final output. This observation has spurred extensive research on layer pruning techniques, which focus on eliminating redundant layers while retaining the model’s core functionalities. LLMDrop~\cite{he2024matters} further discovered that the Attention block is more redundant than the MLP block, highlighting the need for a more fine-grained pruning approach that selectively removes redundant components within each block rather than pruning entire layers. This redundancy provides new insights for optimizing model deployment, enabling more efficient acceleration strategies while maintaining performance.

For both layer and attention pruning, existing methods~\cite{men2024shortgpt,yang-etal-2024-laco,he2024matters,mao2024compressibility} adhere to the \textit{de facto} practice of using cosine similarity to measure the redundancy between computation blocks. Redundant blocks with high similarity scores are identified and removed by comparing adjacent layers or selected layer pairs. However, cosine similarity primarily captures the geometric alignment of hidden representations, which does not necessarily reflect the actual information contribution of each layer. Consequently, relying solely on cosine similarity for pruning may lead to suboptimal decisions, potentially compromising model performance.

In this paper, we reconsider the use of cosine similarity as the criterion for pruning and propose \textbf{EntroDrop}, a novel approach that leverages entropy increase to assess the importance of computation blocks. Empirical analysis reveals that the entropy of hidden representations initially decreases in the early layers but progressively rises across subsequent layers. It suggests that entropy can serve as an effective indicator of information richness within each block. Unlike cosine similarity, which primarily captures geometric relationships, entropy directly quantifies the information content of a block’s output, providing a more reliable basis for pruning decisions. Extensive experiments comparing entropy-based and cosine similarity-based pruning demonstrate that our entropy-driven approach more effectively preserves model accuracy while reducing computational costs. The code is open-sourced to enhance further research~\footnote{https://github.com/SalesforceAIResearch/EntroDrop}. Our key contributions are summarized as:
\begin{itemize}
    \item We conduct an empirical analysis of entropy dynamics in hidden representations across LLM blocks during inference, offering new insights into information flow.
    \item We propose a novel entropy-based pruning strategy that effectively reduces model size while preserving performance.
    \item Extensive experiments demonstrate the superiority of EntroDrop over traditional cosine similarity-based pruning methods.
\end{itemize}

% The remainder of this paper is organized as follows: Section 2 discusses related works to our approach. Section 3 introduces our entropy-based pruning method. Section 4 presents experimental results and analysis. Finally, Section 5 concludes the paper and discusses potential future research directions.

\section{Preliminary}
Transformer-based architectures consist of two primary computational blocks: the Attention and the MLP Block. 
These blocks process input hidden states and enrich them sequentially.

\subsection{Computation Blocks}
\textbf{Attention Block} enables each token in the input sequence to interact with others. Given an input $\mathbf{X}$, a Layer Normalization (LayerNorm) operation is applied before the self-attention computation $\mathbf{X}_{\text{norm}} = \text{LayerNorm}(\mathbf{X})$. Then, the attention mechanism computes as:
\begin{equation}
\mathbf{Y} = \text{Softmax}\left(\frac{\mathbf{Q} \mathbf{K}^T}{\sqrt{d_k}}\right) \mathbf{V},
\end{equation}
% \begin{equation}
% \mathbf{Y} = \mathbf{A} \mathbf{V},
% \end{equation}
where $\mathbf{Q} = \mathbf{X}_{\text{norm}} \mathbf{W}_Q$, $\mathbf{K} = \mathbf{X}_{\text{norm}} \mathbf{W}_K$, $\mathbf{V} = \mathbf{X}_{\text{norm}} \mathbf{W}_V$, and $\sqrt{d_k}$ is a scaling factor. The output $\mathbf{Y}$ represents the transformed hidden states.

\textbf{MLP Block} further transforms the output of Attention block. Assume the input for MLP block is also $\mathbf{X}$. It firstly applies layer normalization to stabilize the output as $\mathbf{X}_{\text{norm}} = \text{LayerNorm}(\mathbf{X})$. Then a two-layer feedforward network is calculated to process $\mathbf{X}_{\text{norm}}$ as:
\begin{equation}
\mathbf{Y} = \text{ReLU}(\mathbf{X}_{\text{norm}}  \mathbf{W}_1 + \mathbf{b}_1) \mathbf{W}_2 + \mathbf{b}_2,
\end{equation}
where $\mathbf{W}_1$, $\mathbf{W}_2$, $\mathbf{b}_1$ and $\mathbf{b}_2$ are learnable parameters. There are also other variants~\cite{touvron2023llama} for this feedforward network. Together, the Attention Block and MLP Block form a complete \textbf{Transformer Block}, which can be stacked to build deep Transformer models. Each Transformer Block refines and enriches the hidden states, enabling hierarchical learning across multiple layers.

\begin{figure*}[ht]
    \begin{center}
    \includegraphics[width=0.3\linewidth]{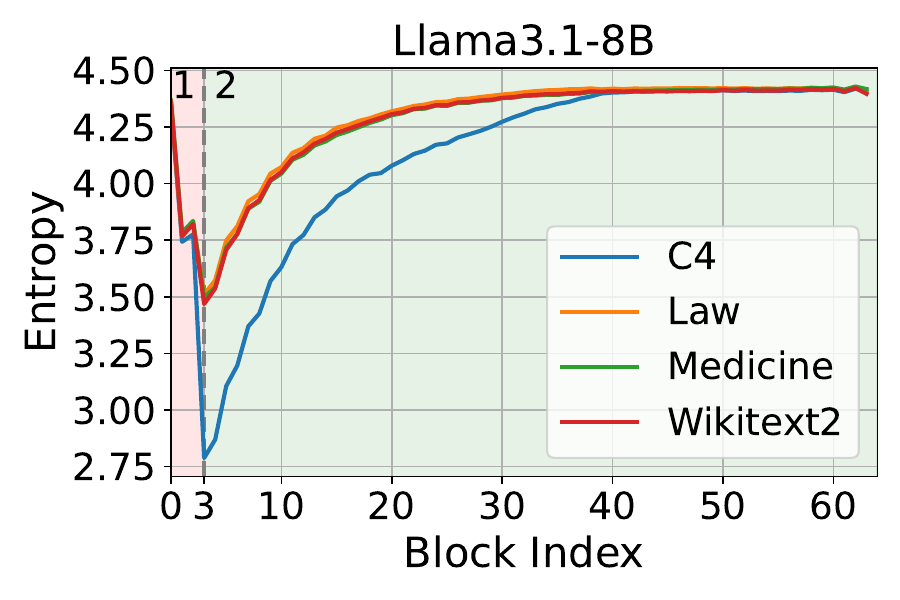}
    \includegraphics[width=0.3\linewidth]{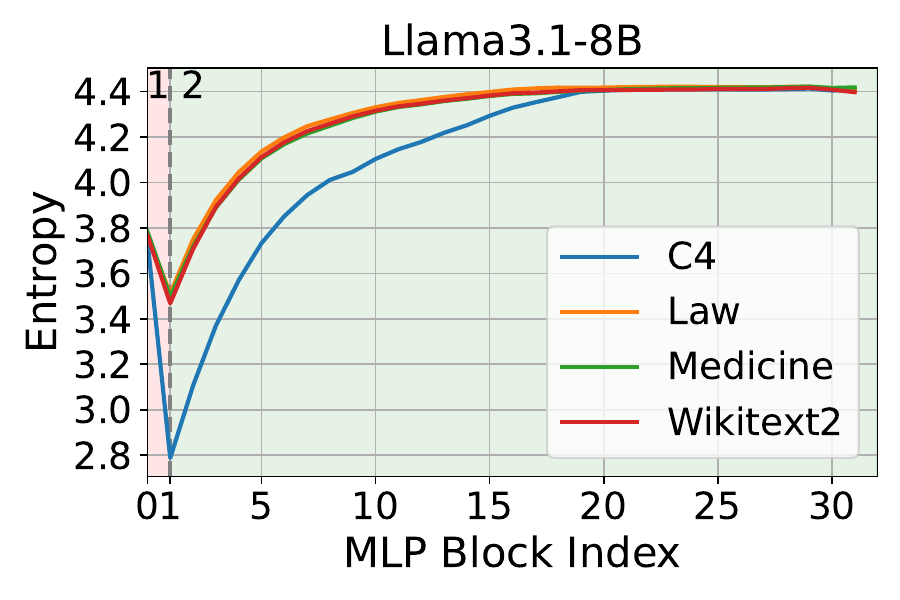}
    \includegraphics[width=0.3\linewidth]{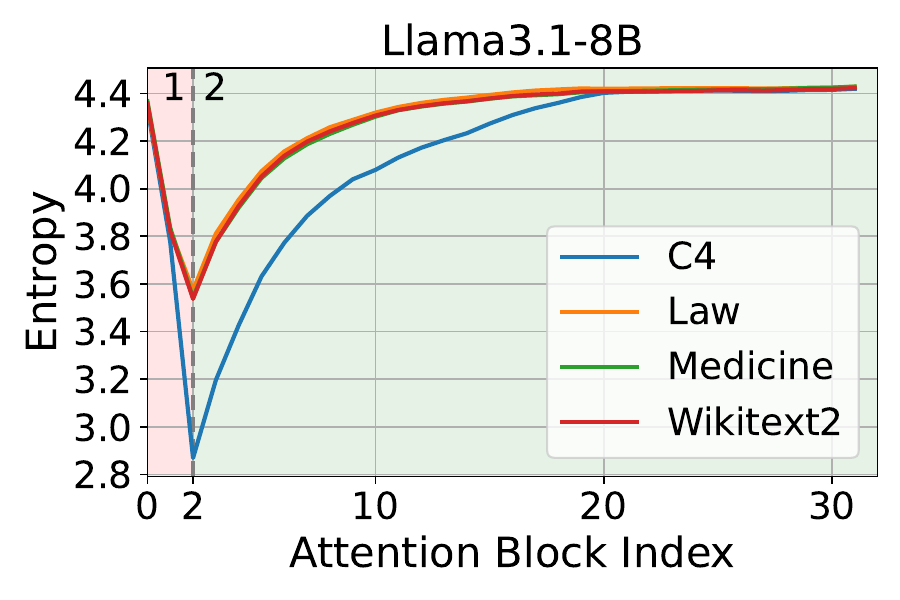}

    \includegraphics[width=0.3\linewidth]{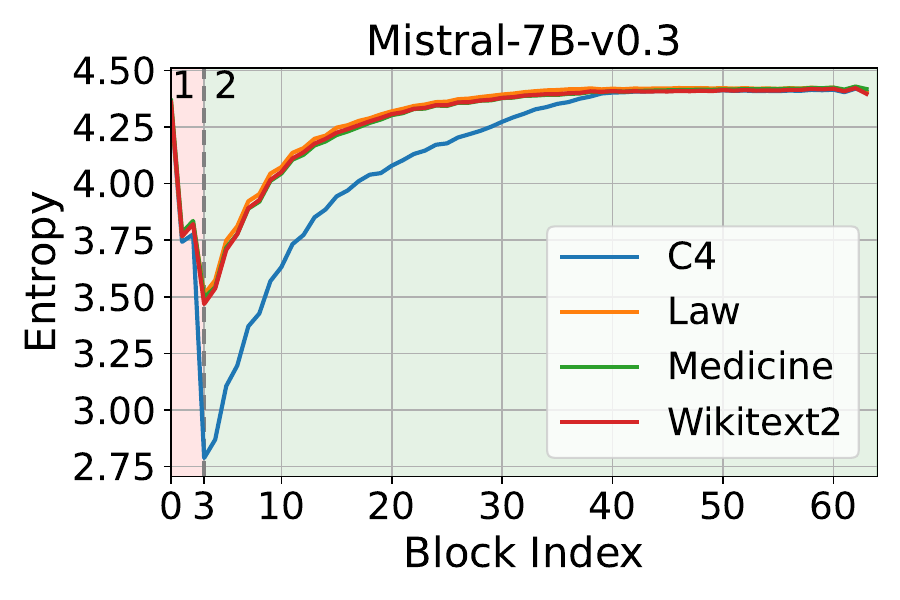}
    \includegraphics[width=0.3\linewidth]{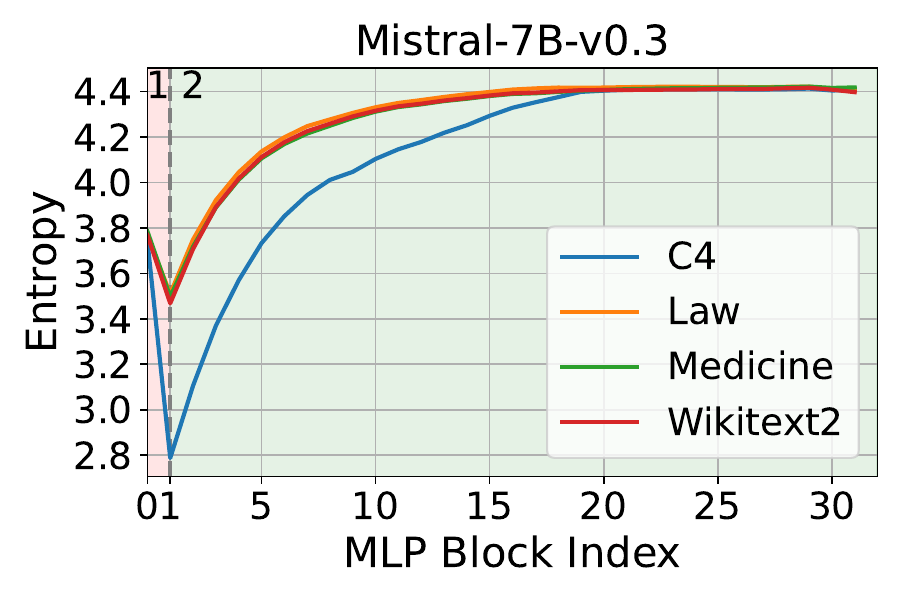}
    \includegraphics[width=0.3\linewidth]{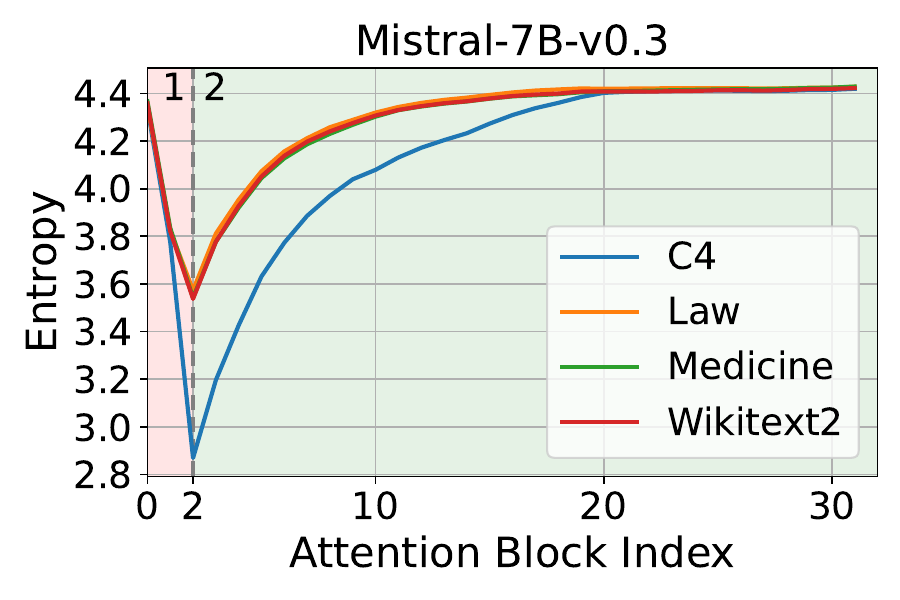}

    \end{center}
    \caption{Entropy Dynamics among Layers during Inference}
    \label{fig:prelim_experiment}
\end{figure*}

\subsection{Block-Wise Pruning}
Block-wise pruning aims to determine the importance of each computation block by analyzing the relationship between its input $\mathbf{X}$ and output $\mathbf{Y}$. The goal is to define an effective metric that identifies less informative blocks for removal while preserving essential model functionality.
To quantify the importance of a block, an importance criterion is often calculated as:
\begin{equation}
I = g(\mathbf{X}, \mathbf{Y})
\end{equation}
where $g(\cdot)$ is a function measuring the information contribution of the block and we prioritize the pruning on blocks with less importance score. No matter on which computation blocks, current methods~\cite{he2024matters,men2024shortgpt} judge the importance by cosine similarity and the importance criterion is calculated as $g(\mathbf{X}, \mathbf{Y}) = 1 - \frac{\mathbf{X} \cdot \mathbf{Y}}{|\mathbf{X}| |\mathbf{Y}|}$. In this paper, we propose entropy increase, a new importance criterion based on empirical observations of entropy change across the layers. Entropy increase can better capture the information flow within the model, providing a more effective metric for identifying redundant blocks.

\subsection{Entropy Estimation}

\section{Method}

\subsection{Observations on Entropy Dynamics}

To investigate the entropy across different layers of Transformer models, we conduct experiments on Llama3.1-8B~\cite{dubey2024llama}~\footnote{https://huggingface.co/meta-llama/Llama-3.1-8B} and Mistral-7B-v0.3~\cite{jiang2023mistral}~\footnote{https://huggingface.co/mistralai/Mistral-7B-v0.3}. We analyze the entropy trends during inference across Transformer Blocks, Attention Blocks, and MLP Blocks using four datasets: C4, Law, Medicine and Wikitext2.
% C4~\cite{raffel2020exploring}~\footnote{\text{https://huggingface.co/datasets/allenai/c4}}, 
% Law~\cite{cheng2023adapting}~\footnote{\text{https://huggingface.co/datasets/AdaptLLM/law\_knowledge\_prob}}, 
% Medicine~\cite{cheng2023adapting}~\footnote{\text{https://huggingface.co/datasets/AdaptLLM/med\_knowledge\_prob}}, and 
% Wikitext2~\cite{merity2016pointer}~\footnote{\text{https://huggingface.co/datasets/mindchain/wikitext2}}. 
We compute the entropy of hidden states at each block level during inference and track its evolution across the entire network. The experimental results, shown in Fig.~\ref{fig:prelim_experiment}, reveal an intriguing phenomenon: the model processes information in two distinct stages during inference:

\begin{itemize}
\item \textbf{Stage 1}: Entropy Decrease (Layers 1–3): In the early layers, entropy progressively decreases, suggesting that the model compresses information, filters redundant features, and forms compact representations.
\item \textbf{Stage 2}: Entropy Increase (Layers 3–32): In the later layers, entropy gradually increases, indicating that the network enriches hidden state representations and expands contextual information.
\end{itemize}

This observation remains consistent across all tested datasets and aligns with previous research~\cite{yang-etal-2024-laco,men2024shortgpt}, which suggests that the initial layers are crucial for information retention and should remain intact. Our empirical findings indicate that these layers play a key role in compressing and structuring input representations. Meanwhile, the entropy increase in later layers supports the idea that they focus on feature expansion rather than compression, making them more suitable candidates for pruning. Furthermore, the gradual entropy increase across these layers suggests that each contributes similarly to transforming hidden states. This insight can be leveraged to design an effective pruning strategy that removes redundant layers while preserving overall model performance.

\subsection{EntroDrop}

\begin{figure}[t]
    \begin{center}
    \includegraphics[width=1.0\linewidth]{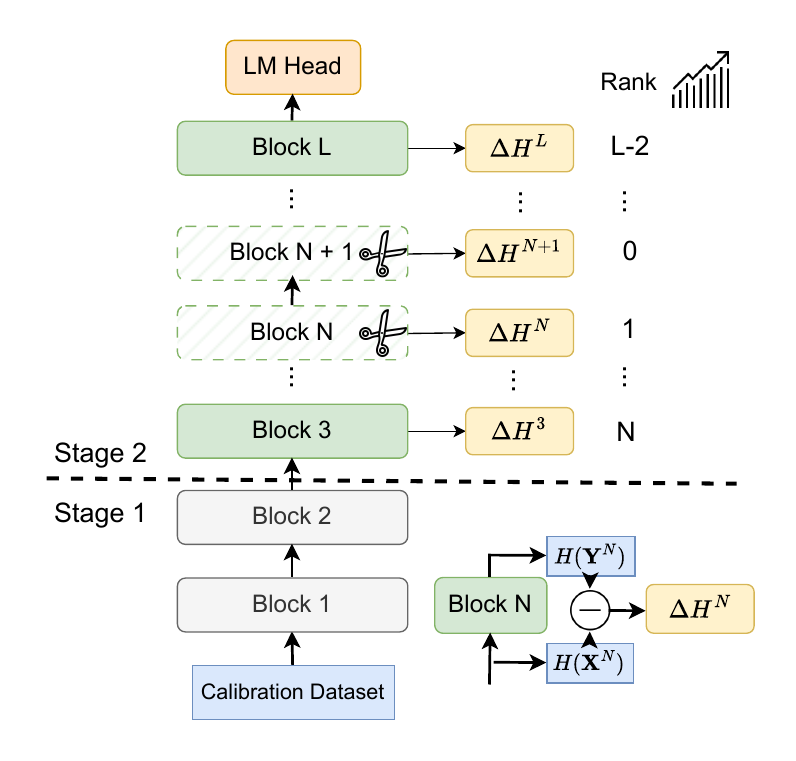}
    \end{center}
    \caption{Overview of the EntroDrop framework. Stage 1 keeps intact, while Stage 2 exhibits increasing entropy. Blocks in Stage 2 are ranked based on their entropy increase, and those with the lowest increase are pruned.}
    \label{fig:framework}
\end{figure}

Based on our empirical observations of entropy dynamics across Transformer models, we propose \textbf{EntroDrop}, a novel entropy-based pruning method that leverages entropy increase in later layers to identify and remove redundant computation blocks while preserving essential model performance. The framework is shown in Fig.~\ref{fig:framework}.

We consider a pre-trained Transformer model consisting of $L$ computation blocks, each responsible for transforming hidden states as the input propagates through the network. Given a calibration dataset $\mathcal{D}$, we pass input samples through the model and collect the hidden states at each block:
\begin{equation}
\mathbf{Z}^{l} = f_l(\mathbf{Z}^{l-1}), \quad l = 1, 2, \dots, L,
\end{equation}
where $\mathbf{Z}^{l}$ represents the hidden state at the $l$-th block, and $f_l(\cdot)$ denotes the computation block function (e.g., Attention, MLP). Once the hidden states at all blocks are obtained, we estimate the entropy of each block and rank them according to their entropy increase values. The lowest $K$ blocks, which exhibit minimal entropy increase, are selected for pruning.

EntroDrop leverages a two-stage pruning strategy based on entropy observations. Stage 1 compresses the information, and no computation blocks are pruned in this stage. Stage 2 gradually increases the entropy, suggesting that these blocks perform similar hidden state enrichments. The transition point between the two stages, denoted as $S_{\text{start}}$, is determined using a calibration dataset.

To effectively estimate the importance of computation blocks, we define entropy increase as:
\begin{equation}
\Delta H^{l} = H(\mathbf{Z}^{l}) - H(\mathbf{Z}^{l-1}),
\end{equation}
where $H(\cdot)$ represents the entropy estimation function. Blocks in Stage 2, indexed by $S_{\text{start}} \leq l \leq L$, are ranked in ascending order based on their entropy increase:
\begin{equation}
\operatorname{Rank}(\Delta H^{l}) = \operatorname{argsort}(\Delta H^{l}) \quad \text{for } l \geq S_{\text{start}}.
\end{equation}
Finally, the $K$ blocks with the smallest entropy increase within Stage 2 are selected for pruning:
\begin{equation}
\mathcal{S}_{\text{prune}} = \{ f_{i} \mid f_{i} \in \operatorname{Rank}(\Delta H^{l})_{S_{\text{start}}:L}[:K] \},
\end{equation}
where $\mathcal{S}_{\text{prune}}$ denotes the set of pruned blocks and $\Delta H^{l}$ represents the entropy increase of $l$ computation block.
The bottom $k$ ranked layers are pruned to optimize efficiency.
To estimate entropy efficiently, we explore multiple estimation techniques: 
\begin{itemize}
    \item Bucket-based Estimation: Discretizes activation values into bins and estimates entropy based on frequency distribution.
    \item  K-Nearest Neighbors (KNN): Computes entropy by estimating local density using KNN. 
    \item Renyi Entropy: A generalization of Shannon entropy that provides a tunable parameter to control sensitivity to distribution variations.
\end{itemize}

Regardless of the estimation method used, entropy computation remains efficient, making EntroDrop a practical pruning strategy. Our experimental results demonstrate that selecting an appropriate entropy estimation method is crucial for achieving optimal pruning performance. Among the approaches tested, Bucket-based estimation and KNN-based estimation were found to be particularly effective in preserving model accuracy.

\begin{table*}[t]
\centering
\setlength{\tabcolsep}{3pt} % 调整列间距
\renewcommand{\arraystretch}{1.2} % 调整行距，默认为 1
\small
\resizebox{\textwidth}{!}{ % 让表格适应页面宽度
\begin{tabular}{c|c|ccccccccccc|c}
\hline
\multirow{2}{*}{\textbf{L}} & \multirow{2}{*}{\textbf{Method}} & \multicolumn{11}{|c|}{\textbf{Dataset}} & \multirow{2}{*}{\textbf{Average}} \\
\cline{3-13}
 &  & PIQA & HellaSwag & WSC273 & CSQA & WinoGrande & ARC-E & ARC-C & OBQA & MMLU & CMMLU & RACE &  \\
\hline
0 & * & 0.7998 & 0.6003 & 0.8608 & 0.7166 & 0.7316 & 0.8148 & 0.5102 & 0.334 & 0.6332 & 0.509 & 0.3923 & 0.6275 \\\hline \hline
\multirow{5}{*}{4} & LaCo & 0.7628 & 0.5116 & 0.8059 & 0.6806 & 0.7103 & 0.7302 & 0.4462 & 0.284 & 0.5949 & 0.437 & 0.3761 & 0.5763 \\
 & ShortGPT & 0.7557 & 0.5504 & 0.7949 & 0.6921 & 0.7017 & 0.7222 & 0.442 & 0.312 & 0.5802 & 0.416 & 0.3818 & 0.5772 \\
 & Ours (Layer) & 0.7573 & 0.5407 & 0.8242 & 0.7027 & 0.7088 & 0.7504 & 0.4275 & 0.286 & 0.6212 & 0.4918 & 0.3818 & \textbf{0.5902} \\ \cline{2-14}
 
 & LLMDrop & 0.8025 & 0.5965 & 0.8352 & 0.7117 & 0.7498 & 0.8194 & 0.5188 & 0.342 & 0.6312 & 0.5111 & 0.3933 & \textbf{0.6283} \\
 & Ours (Attn) & 0.8003 & 0.6022 & 0.8498 & 0.7158 & 0.7364 & 0.8157 & 0.5179 & 0.342 & 0.6238 & 0.5044 & 0.3895 & 0.6271 \\\hline \hline

\multirow{5}{*}{8} & LaCo & 0.6197 & 0.3098 & 0.6007 & 0.4005 & 0.6227 & 0.3952 & 0.2756 & 0.236 & 0.4463 & 0.3405 & 0.2478 & 0.4086 \\
 & ShortGPT & 0.6045 & 0.2825 & 0.5971 & 0.4046 & 0.5422 & 0.4289 & 0.2739 & 0.182 & 0.3226 & 0.3153 & 0.2526 & 0.3824 \\
  & Ours (Layer) & 0.6795 & 0.4384 & 0.7509 & 0.6216 & 0.6898 & 0.5644 & 0.3532 & 0.212 & 0.5584 & 0.4408 & 0.3378 & \textbf{0.5133} \\ \cline{2-14}
  
 & LLMDrop & 0.7954 & 0.5877 & 0.8388 & 0.7174 & 0.7443 & 0.8119 & 0.5068 & 0.356 & 0.6338 & 0.5073 & 0.401 & \textbf{0.6273} \\
 & Ours (Attn) & 0.7954 & 0.5921 & 0.8352 & 0.7183 & 0.7411 & 0.8186 & 0.5154 & 0.354 & 0.6301 & 0.5041 & 0.3876 & 0.6265 \\\hline \hline

\multirow{5}{*}{12} & LaCo & 0.6202 & 0.3312 & 0.6337 & 0.1966 & 0.6219 & 0.4293 & 0.2705 & 0.198 & 0.2428 & 0.2571 & 0.2813 & 0.3711 \\
 & ShortGPT & 0.6007 & 0.3066 & 0.5861 & 0.516 & 0.5501 & 0.4007 & 0.2765 & 0.178 & 0.3605 & 0.3252 & 0.266 & \textbf{0.3969} \\
 & Ours (Layer) & 0.6007 & 0.3066 & 0.5861 & 0.516 & 0.5501 & 0.4007 & 0.2765 & 0.178 & 0.3605 & 0.3252 & 0.266 & \textbf{0.3969}  \\ \cline{2-14}
 
 & LLMDrop & 0.7867 & 0.5584 & 0.8608 & 0.679 & 0.7253 & 0.7807 & 0.4753 & 0.31 & 0.5992 & 0.4511 & 0.3799 & \textbf{0.6006} \\
 & Ours (Attn) & 0.7867 & 0.5584 & 0.8608 & 0.679 & 0.7253 & 0.7807 & 0.4753 & 0.31 & 0.5992 & 0.4511 & 0.3799 & \textbf{0.6006} \\\hline \hline

\multirow{5}{*}{16} & LaCo & 0.5854 & 0.2904 & 0.6447 & 0.1957 & 0.5612 & 0.3443 & 0.2338 & 0.16 & 0.2295 & 0.2527 & 0.2469 & \textbf{0.3404} \\
 & ShortGPT & 0.5647 & 0.2754 & 0.5421 & 0.1949 & 0.5501 & 0.3194 & 0.244 & 0.154 & 0.2295 & 0.2529 & 0.2488 & 0.3251 \\
  & Ours (Layer) & 0.5729 & 0.2705 & 0.5238 & 0.2113 & 0.5099 & 0.3165 & 0.2321 & 0.138 & 0.2627 & 0.2538 & 0.2278 & 0.3199 \\ \cline{2-14}
  
 & LLMDrop & 0.6926 & 0.4272 & 0.7875 & 0.2121 & 0.7017 & 0.564 & 0.3328 & 0.222 & 0.2735 & 0.2819 & 0.2938 & 0.4354 \\
 & Ours (Attn) & 0.7514 & 0.4481 & 0.7656 & 0.3022 & 0.7048 & 0.6595 & 0.3925 & 0.27 & 0.3586 & 0.2784 & 0.3282 & \textbf{0.4781} \\\hline \hline
\end{tabular}
}
\caption{Experiment Results on Llama3.1-8B. The best performance is marked in bold.}
\label{tab:main_llama}
\end{table*}

\section{Experiments}
In this section, we conduct comprehensive experiments to evaluate the effectiveness of \textbf{EntroDrop} from different perspectives. 

% Specifically, we aim to address the following research questions (RQ):

% \begin{itemize}
%     \item \textbf{RQ1}: How does EntroDrop compare with other block pruning methods?
%     \item \textbf{RQ2}: What is the impact of the calibration dataset on pruning performance?
%     \item \textbf{RQ3}: How sensitive is EntroDrop to different entropy estimation methods?
%     \item \textbf{RQ4}: What is the actual speedup achieved after pruning blocks?
% \end{itemize}

\subsection{Experimental Setup}

\textbf{Models} We conduct experiments on two state-of-the-art decoder-only Transformer models: Llama3.1-8B and Mistral-7B-v0.3. To make a fair comparison, all experiments are finished on a single 40G A100 GPU device.

\textbf{Benchmarks}
To evaluate the effectiveness of \textbf{EntroDrop}, we test on a diverse set of reasoning and comprehension benchmarks:
Commonsense Reasoning: PIQA~\cite{Bisk2020}, HellaSwag~\cite{zellers2019hellaswag}, WSC273~\cite{sakaguchi2021winogrande}, CSQA~\cite{talmor-etal-2019-commonsenseqa}, WinoGrande~\cite{sakaguchi2021winogrande}.
Scientific and Knowledge-based QA: ARC-E~\cite{clark2018think}, ARC-C~\cite{clark2018think}, OBQA~\cite{DBLP:conf/emnlp/MihaylovCKS18}.
General and Subject-specific Knowledge: MMLU~\cite{hendryckstest2021,hendrycks2021ethics}, CMMLU~\cite{DBLP:conf/acl/0002ZKY0GDB24}, RACE~\cite{lai-etal-2017-race}.
These benchmarks cover a wide range of language abilities, from commonsense understanding to complex multi-choice question answering.

\textbf{Baselines} We compare \textbf{EntroDrop} against state-of-the-art pruning techniques in two categories:
(1) Layer Pruning Methods that directly prune the whole transformer block:
LaCo~\cite{yang-etal-2024-laco} and ShortGPT~\cite{men2024shortgpt}.
(2) Attention Pruning Method that only prunes the attention block: LLMDrop~\cite{he2024matters}.
These baselines allow us to assess how EntroDrop compares against existing pruning methods in terms of performance preservation under different pruning granularity.

\subsection{Overall Performance}

\begin{table*}[t]
\centering
\setlength{\tabcolsep}{3pt} % Adjust column spacing
\renewcommand{\arraystretch}{1.2} % Adjust row spacing, default is 1
\small
\resizebox{\textwidth}{!}{ % 让表格适应页面宽度
\begin{tabular}{c|c|ccccccccccc|c}
\hline
\multirow{2}{*}{\textbf{L}} & \multirow{2}{*}{\textbf{Method}} & \multicolumn{11}{|c|}{\textbf{Dataset}} & \multirow{2}{*}{\textbf{Average}} \\
\cline{3-13}
 &  & PIQA & HellaSwag & WSC273 & CSQA & WinoGrande & ARC-E & ARC-C & OBQA & MMLU & CMMLU & RACE &  \\
\hline 
0 & * & 0.803 & 0.6091 & 0.8864 & 0.5741 & 0.7388 & 0.7963 & 0.4898 & 0.33 & 0.5908 & 0.383 & 0.4086 & 0.6009 \\
\hline \hline
\multirow{5}{*}{4} & LaCo & 0.5501 & 0.2975 & 0.6996 & 0.1196 & 0.6275 & 0.2908 & 0.2568 & 0.2400 & 0.1817 & 0.2070 & 0.2746 & 0.3405 \\
 & ShortGPT & 0.7557 & 0.5458 & 0.8352 & 0.4808 & 0.7048 & 0.7104 & 0.4181 & 0.2620 & 0.4887 & 0.3598 & 0.3828 & 0.5404 \\
& Ours (Layer) & 0.7524 & 0.5467 & 0.8278 & 0.4865 & 0.7214 & 0.7079 & 0.4266 & 0.2700 & 0.4954 & 0.3458 & 0.3914 & \textbf{0.5429} \\ \cline{2-14}

 & LLMDrop & 0.8047 & 0.6051 & 0.8791 & 0.5717 & 0.7285 & 0.7971 & 0.4872 & 0.338 & 0.5898 & 0.3828 & 0.3962 & 0.5982 \\
 & Ours (Attn) & 0.802 & 0.6062 & 0.8755 & 0.5725 & 0.7309 & 0.7984 & 0.4889 & 0.338 & 0.5888 & 0.3819 & 0.4019 & \textbf{0.5986}   \\
\hline \hline

\multirow{5}{*}{8} & LaCo & 0.5952 & 0.3180 & 0.7033 & 0.2080 & 0.6377 & 0.3914 & 0.3029 & 0.1940 & 0.2802 & 0.2602 & 0.3062 & 0.3816 \\
& ShortGPT & 0.6627 & 0.3960 & 0.7143 & 0.5184 & 0.6598 & 0.5025 & 0.3294 & 0.2100 & 0.5086 & 0.3259 & 0.3167 & \textbf{0.4677} \\
& Ours (Layer) & 0.6627 & 0.3960 & 0.7143 & 0.5184 & 0.6598 & 0.5025 & 0.3294 & 0.2100 & 0.5086 & 0.3259 & 0.3167 & \textbf{0.4677} \\ \cline{2-14}

 & LLMDrop & 0.7998 & 0.597 & 0.8718 & 0.5766 & 0.7364 & 0.7934 & 0.4753 & 0.332 & 0.5917 & 0.3659 & 0.3952 & 0.5941   \\
 & Ours & 0.8003 & 0.5991 & 0.8681 & 0.5782 & 0.7332 & 0.795 & 0.4855 & 0.324 & 0.5902 & 0.3752 & 0.3952 & \textbf{0.5949}   \\
\hline \hline

\multirow{5}{*}{12} & LaCo & 0.5724 & 0.2937 & 0.6410 & 0.1630 & 0.5825 & 0.3013 & 0.2671 & 0.2020 & 0.2810 & 0.2214 & 0.2584  & 0.3440 \\
& ShortGPT & 0.5702 & 0.2795 & 0.6007 & 0.1974 & 0.5612 & 0.3367 & 0.2858 & 0.2080 & 0.2264 & 0.2426 & 0.2287 & 0.3397 \\
& Ours & 0.6066 & 0.3415 & 0.6154 & 0.2424 & 0.5770 & 0.4146 & 0.2969 & 0.1820 & 0.3169 & 0.2595 & 0.3024 & \textbf{0.3777} \\ \cline{2-14}

 & LLMDrop & 0.7742 & 0.5614 & 0.8388 & 0.4054 & 0.7277 & 0.7483 & 0.4437 & 0.282 & 0.5551 & 0.3143 & 0.3722 & 0.5476   \\
 & Ours & 0.7802 & 0.5749 & 0.8498 & 0.5446 & 0.7222 & 0.7546 & 0.4693 & 0.308 & 0.5857 & 0.3636 & 0.3799 & \textbf{0.5757}   \\
\hline \hline

\multirow{5}{*}{16} & LaCo & 0.5577 & 0.2764 & 0.5165 & 0.2146 & 0.5367 & 0.3266 & 0.2509 & 0.1520 & 0.2637 & 0.2549 & 0.2641 & \textbf{0.3286} \\
& ShortGPT & 0.5403 & 0.2704 & 0.5824 & 0.2031 & 0.5272 & 0.3068 & 0.2619 & 0.1580 & 0.2367 & 0.2539 & 0.2287 & 0.3245 \\
& Ours & 0.5272 & 0.2760 & 0.5275 & 0.1900 & 0.5067 & 0.2955 & 0.2491 & 0.1720 & 0.2473 & 0.2509 & 0.2411 & 0.3167 \\
 \cline{2-14}
 
 & LLMDrop & 0.722 & 0.4572 & 0.8059 & 0.1974 & 0.6946 & 0.5812 & 0.3677 & 0.24 & 0.2525 & 0.2568 & 0.3206 & 0.4451   \\
 & Ours & 0.7497 & 0.4983 & 0.8168 & 0.3178 & 0.7024 & 0.6679 & 0.3968 & 0.244 & 0.4198 & 0.2892 & 0.3407 & \textbf{0.4949}   \\
\hline \hline
\end{tabular}
}
\caption{Experiment Results on Mistral-7B-v0.3. The best performance is marked in bold.}
\label{tab:main_mistral}
\end{table*}

Our experimental results on Llama3.1-8B (Table~\ref{tab:main_llama}) and Mistral-7B-v0.3 (Table~\ref{tab:main_mistral}) demonstrate the effectiveness of \textbf{EntroDrop}. We summarize the key findings as follows:

\begin{itemize}
    \item \textbf{EntroDrop is effective across multiple models.}  
    Our method consistently achieves the best performance across both Llama3.1-8B and Mistral-7B-v0.3. This demonstrates that \textbf{EntroDrop} is a generalizable pruning strategy applicable to different pre-trained LLMs.

    \item \textbf{EntroDrop outperforms both layer pruning and attention pruning baselines.}  
    Compared to LaCo and ShortGPT (layer pruning) and LLMDrop (attention pruning), our method consistently achieves superior results. This suggests that our entropy-based metric effectively identifies and prunes redundant computation blocks at different granularities.

    \item \textbf{Pretrained Transformer models contain significant redundancy, especially in attention layers.}  
    Our experiments show that removing up to 12 layers (37.5\% of total attention layers) in Llama3.1-8B still retains over 95\% of the model’s original performance. This indicates that modern Transformers are often over-parameterized and that structured pruning can significantly improve efficiency without major performance degradation.
\end{itemize}
Overall, these findings confirm that entropy-based pruning is an effective and generalizable strategy for reducing redundant computation in large Transformer models. By leveraging entropy dynamics, \textbf{EntroDrop} enables efficient pruning while maintaining competitive performance across diverse tasks and architectures.

\subsection{Impact of Calibration Dataset}

\begin{figure}[t]
    \begin{center}
    \includegraphics[width=0.9\linewidth]{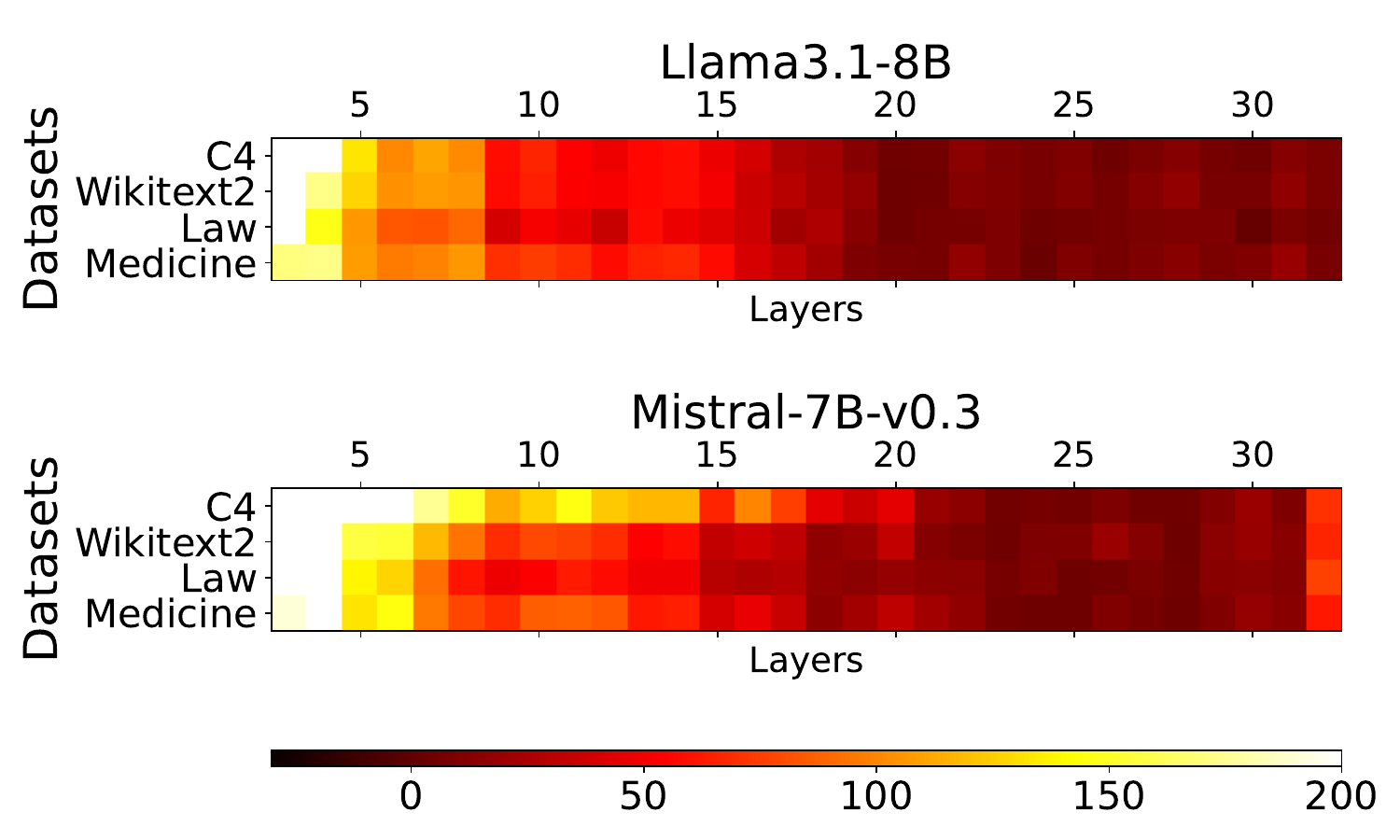}
    \end{center}
    \caption{Heatmap of Calibration Datasets}
    \label{fig:calibration_datasets_heatmap}
\end{figure}

\begin{figure}[t]
    \begin{center}
    \includegraphics[width=0.95\linewidth]{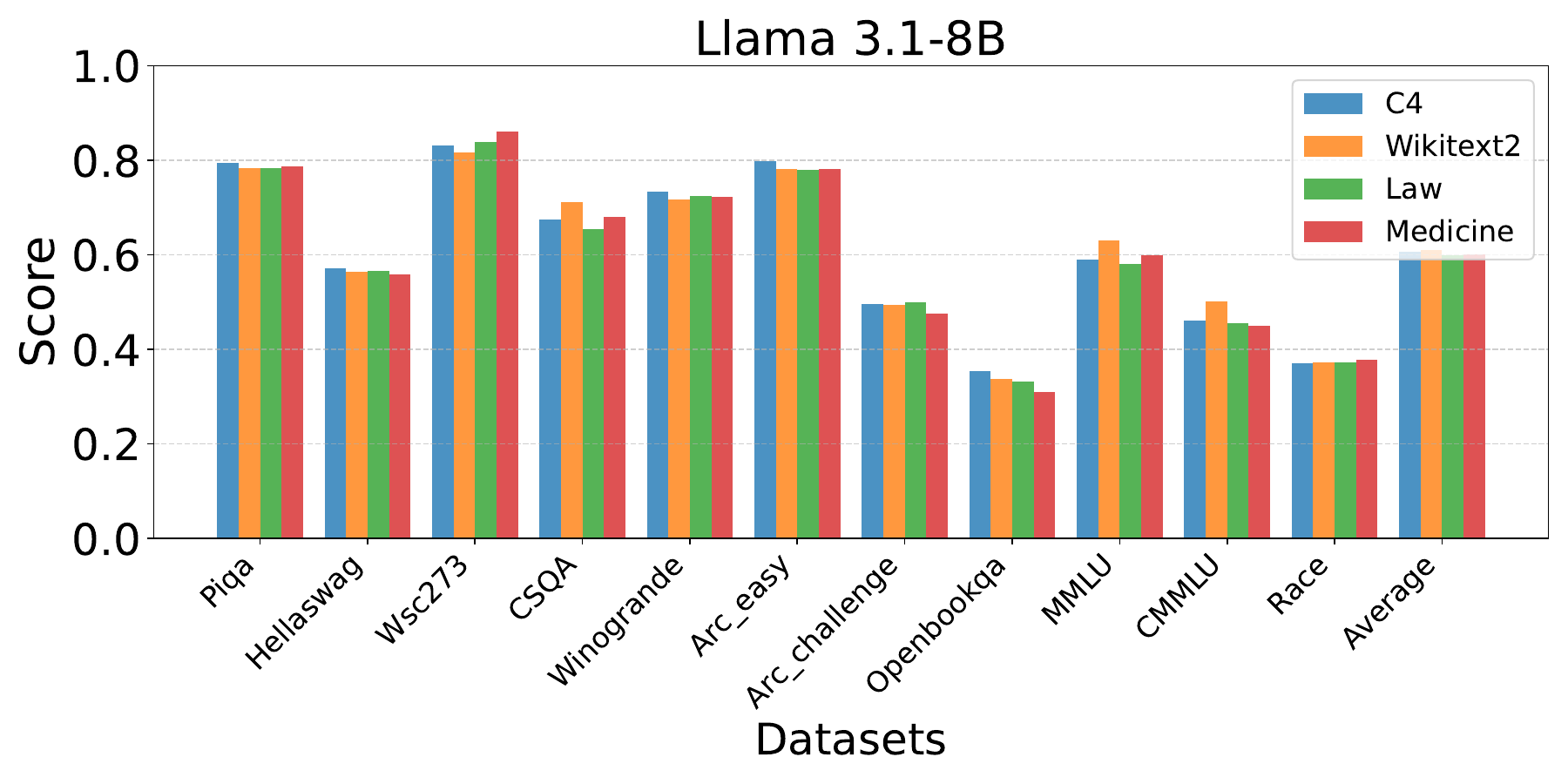}
    \includegraphics[width=0.95\linewidth]{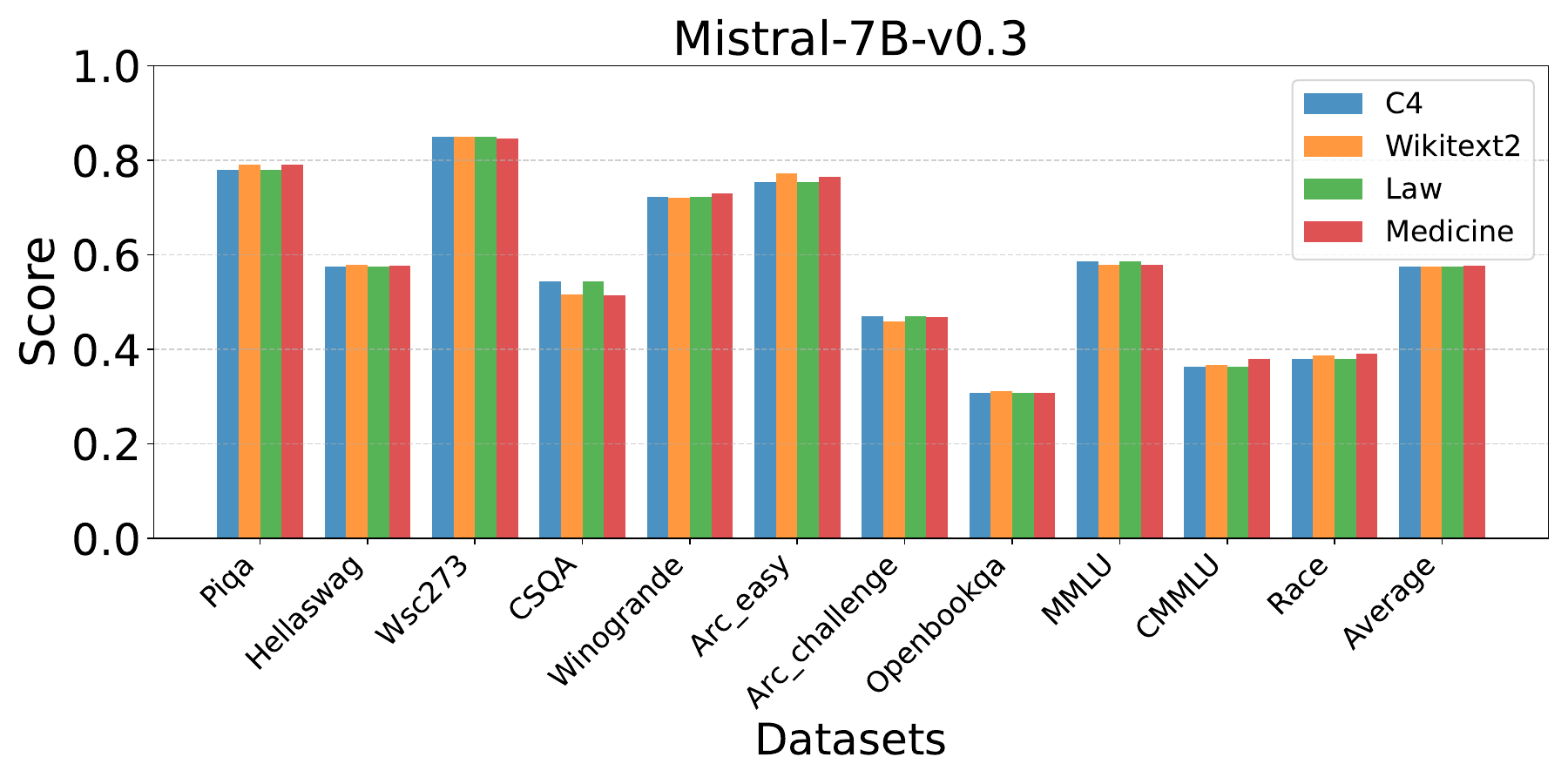}
    \end{center}
    \caption{Impact of Calibration Datasets.}
    \label{fig:calibration_datasets}
\end{figure}

Our pruning method relies on a \textbf{calibration dataset} to estimate entropy dynamics across Transformer layers. We investigate how different calibration datasets affect pruning results. Specifically, we evaluate two general-domain datasets (C4~\cite{raffel2020exploring} and Wikitext~\cite{merity2016pointer}) and two specific-domain datasets (Medicine~\cite{cheng2023adapting} and Law~\cite{cheng2023adapting}).

Figure~\ref{fig:calibration_datasets_heatmap} presents the entropy increase heatmaps estimated using different calibration datasets on Llama3.1-8B and Mistral-7B-v0.3. Across all models, entropy increase is smaller in deeper layers, indicating that these layers contribute less to new information processing and are more redundant. This suggests that deeper layers are natural candidates for pruning. Furthermore, despite differences in calibration datasets, the estimated entropy increase trends remain largely consistent. The relative importance of layers is preserved across general and domain-specific datasets, suggesting that our entropy-based pruning approach is robust to calibration dataset variations. 

To further examine the impact of calibration datasets on model performance, Figure~\ref{fig:calibration_datasets} presents the evaluation results of Llama3.1-8B and Mistral-7B-v0.3 after pruning 12 attention layers (37.5\%). The results show that different calibration datasets lead to minimal differences in performance across all benchmark datasets, reinforcing the robustness of our entropy-based pruning strategy. Notably, even with domain-specific datasets (Medicine, Law), the average accuracy remains stable, indicating that the entropy estimation process generalizes well across different calibration datasets.
These findings confirm that \textbf{EntroDrop} remains effective regardless of the calibration dataset, making it a flexible and generalizable pruning strategy.

\subsection{Entropy Estimation Sensitivity}

\begin{figure}[t]
    \begin{center}
    \includegraphics[width=0.95\linewidth]{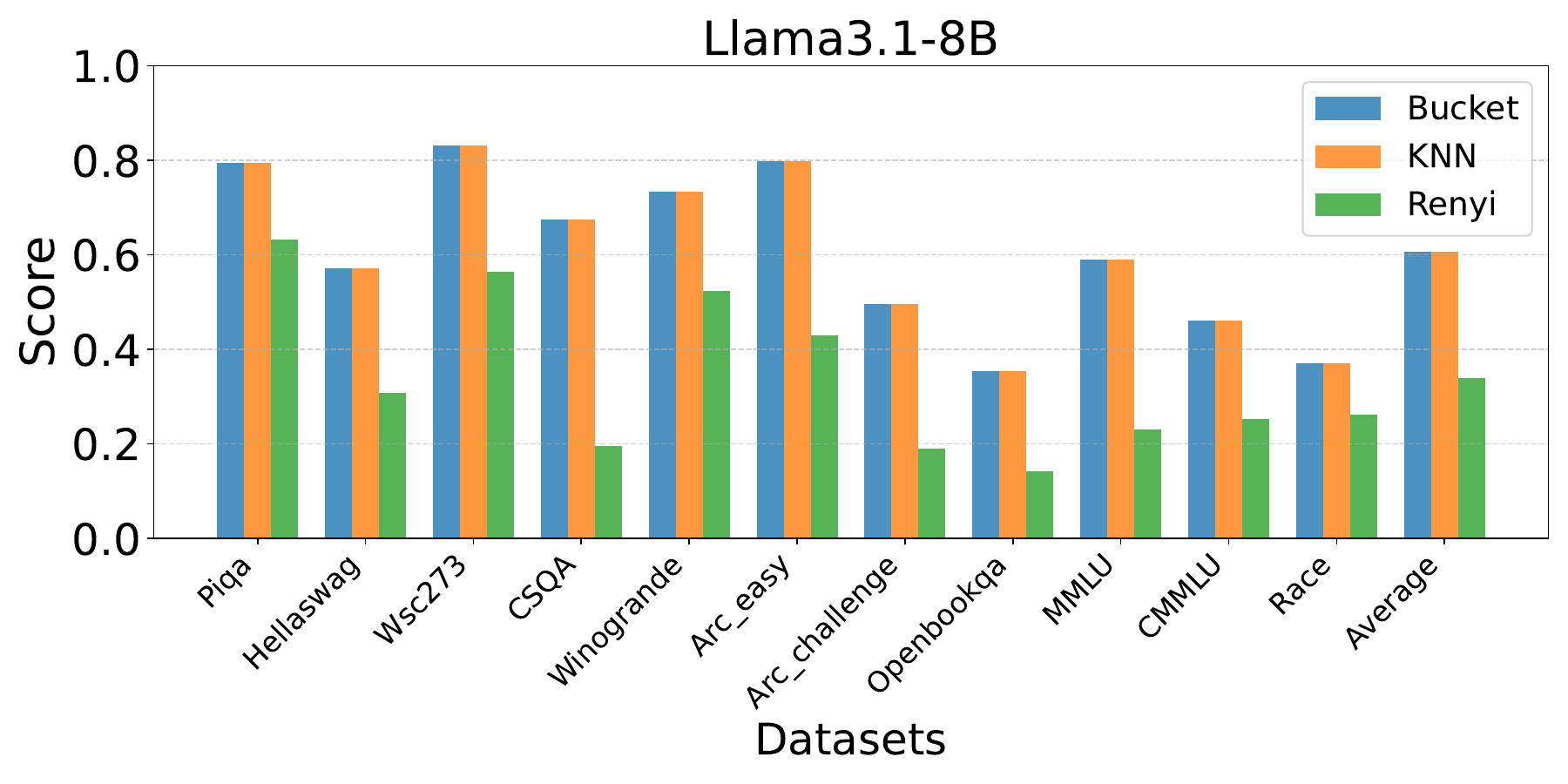}
    \includegraphics[width=0.95\linewidth]{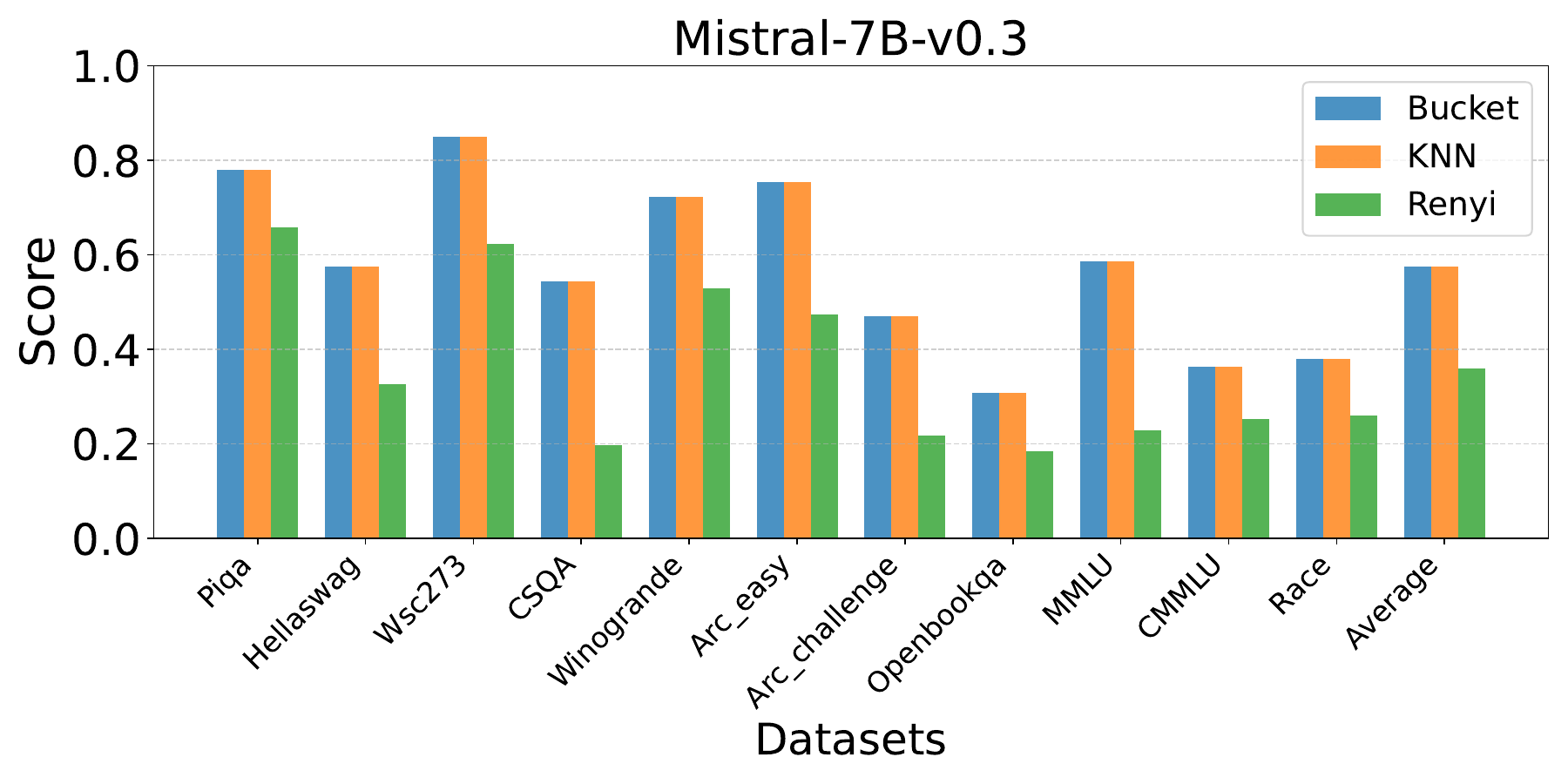}
    \end{center}
    \caption{Impact of Entropy Estimate Methods.}
    \label{fig:entropy_estimation}
\end{figure}

\begin{figure*}[ht]
    \begin{center}
    \includegraphics[width=.245\linewidth]{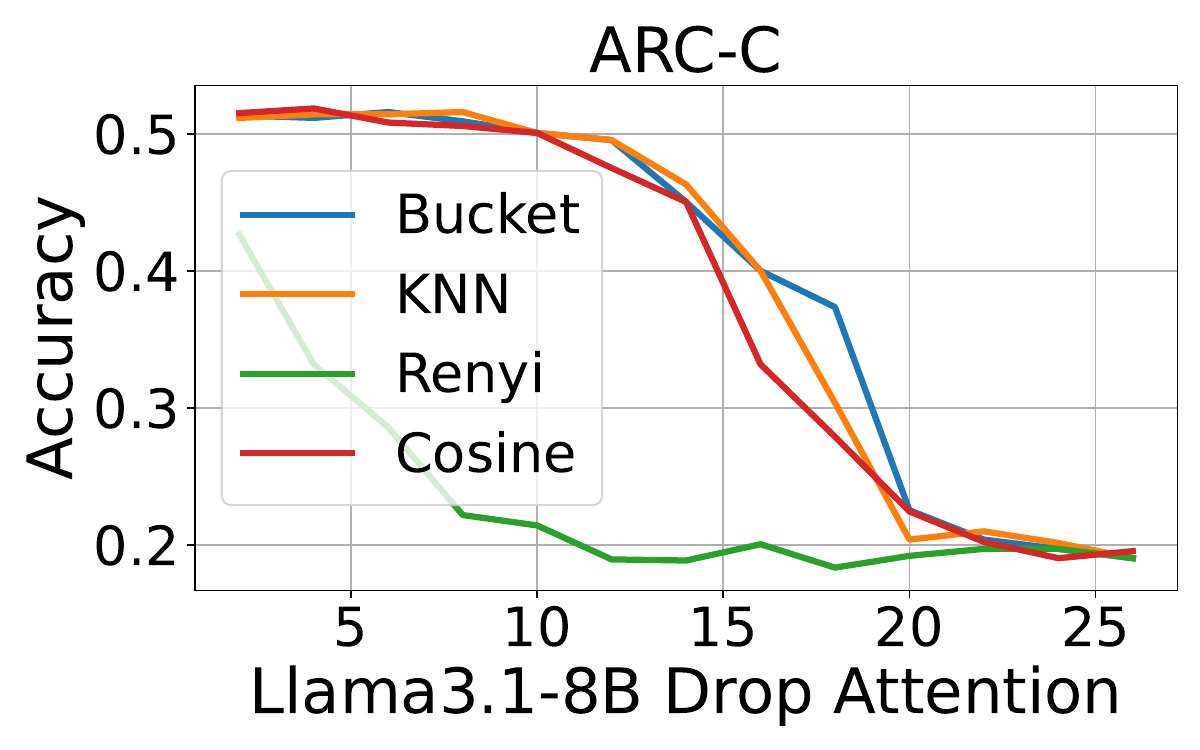}
    \includegraphics[width=.245\linewidth]{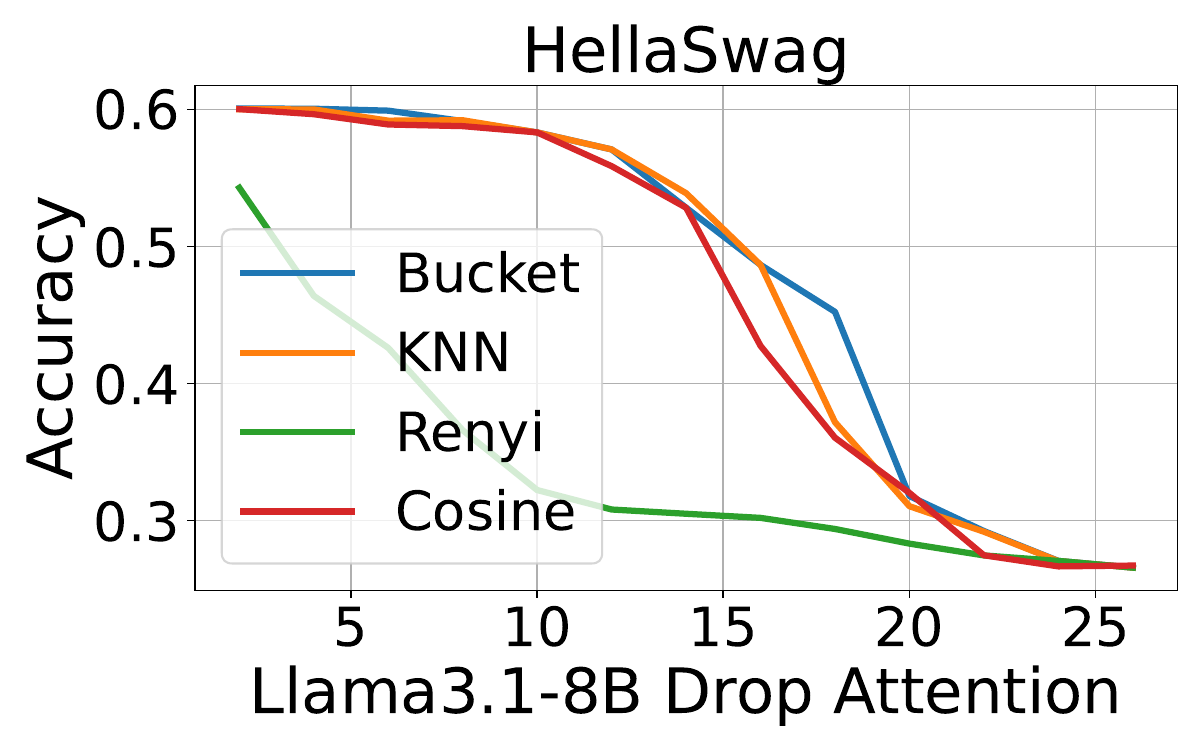}
    \includegraphics[width=.245\linewidth]{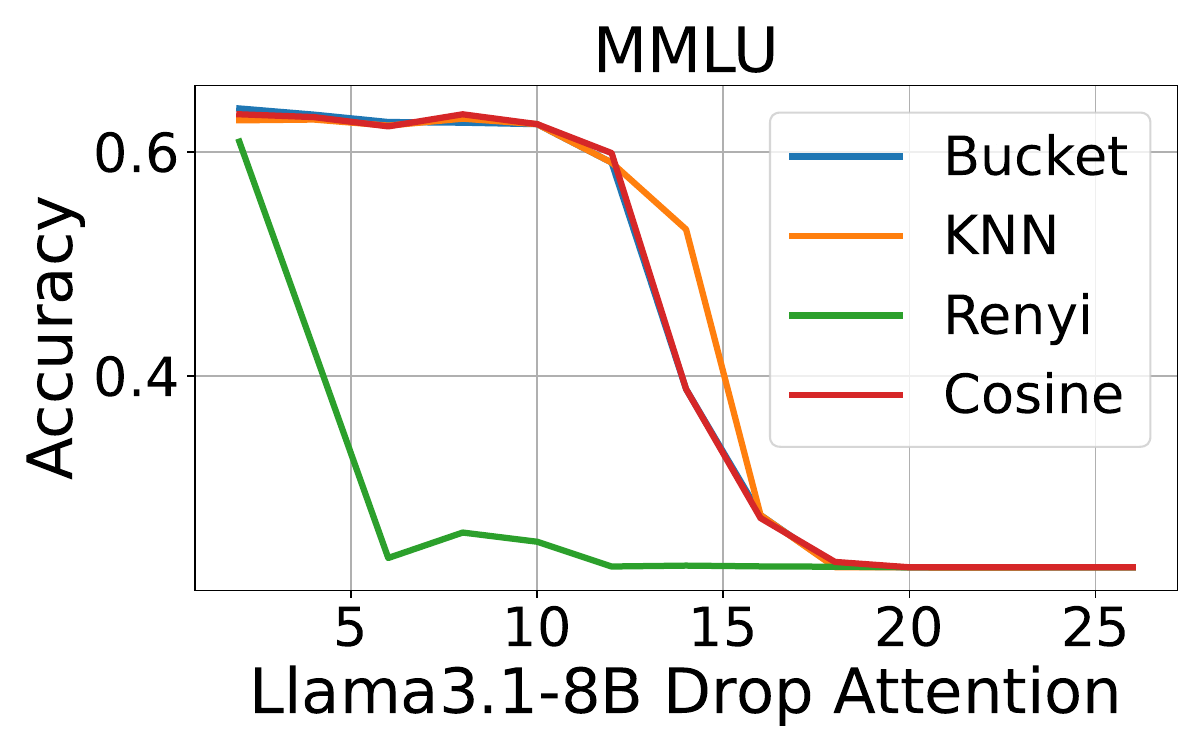}
    \includegraphics[width=.245\linewidth]{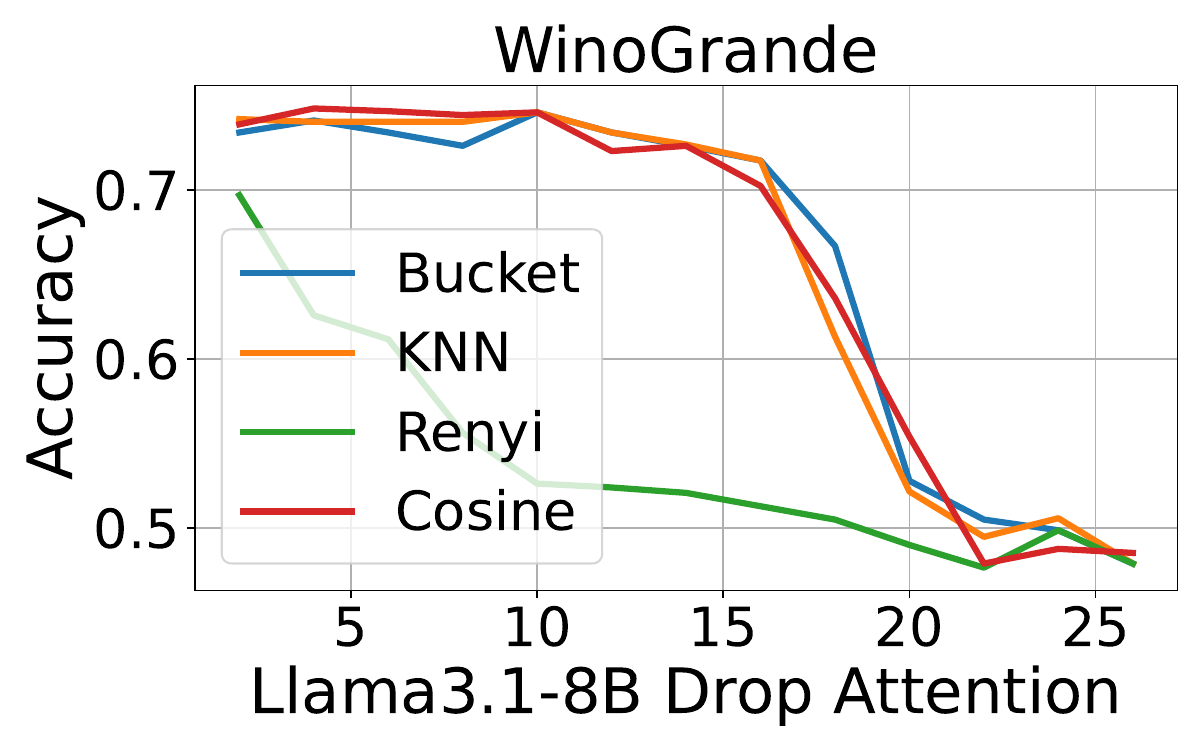}

    \includegraphics[width=.245\linewidth]{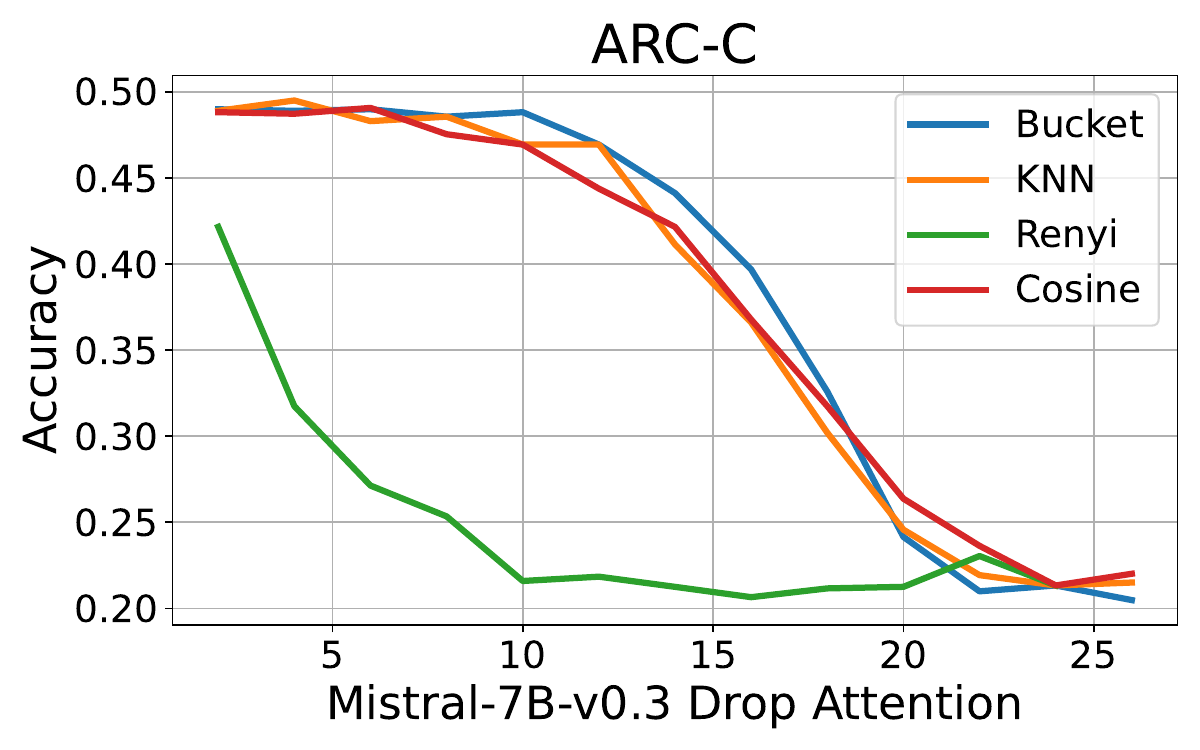}
    \includegraphics[width=.245\linewidth]{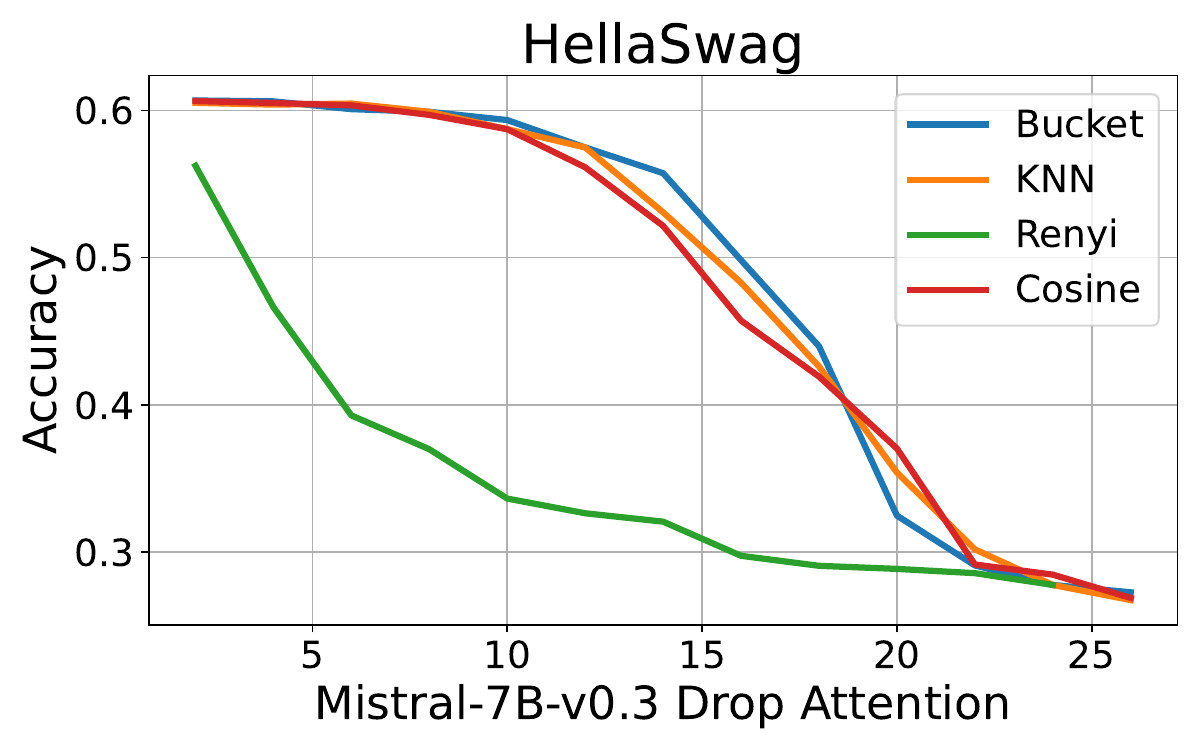}
    \includegraphics[width=.245\linewidth]{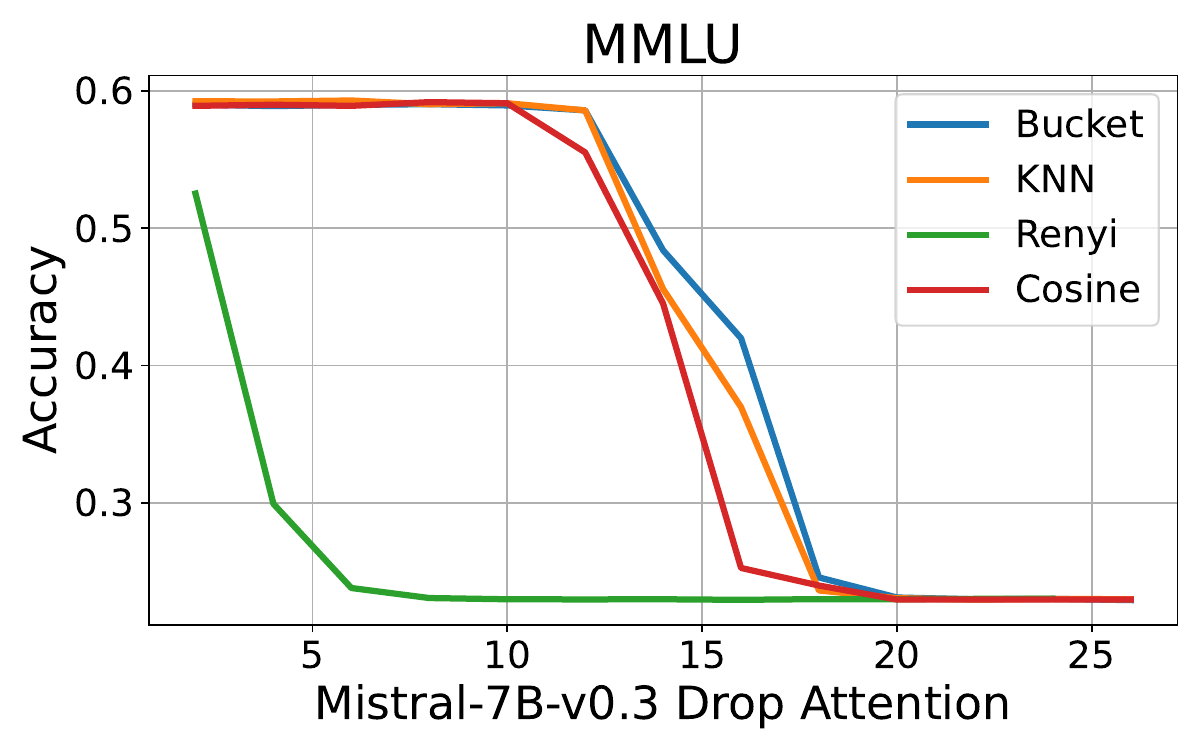}
    \includegraphics[width=.245\linewidth]{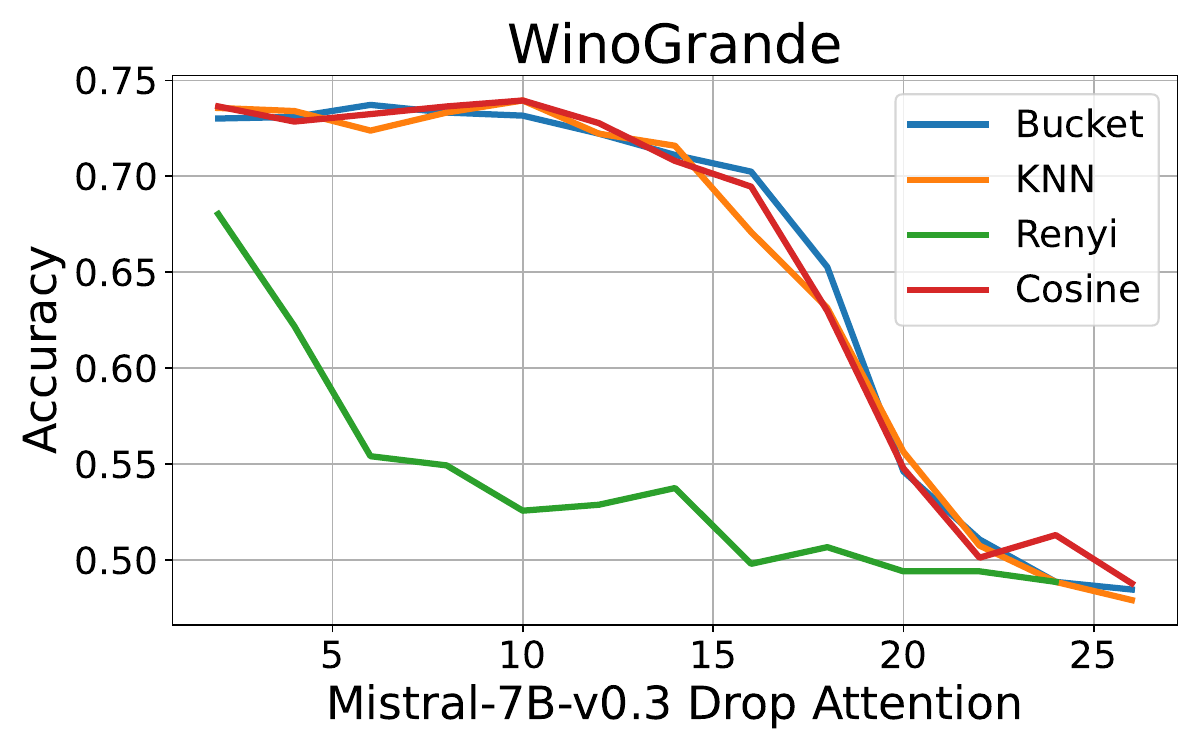}

    \end{center}
    \caption{Attention Deletion Experiments}
    \label{fig:layer_deletion}
\end{figure*}

\textbf{Estimation Method}. Entropy estimation plays a crucial role in our pruning framework, as it directly influences the selection of redundant computation blocks. We evaluate three entropy estimation methods: Bucket, KNN and Renyi. To analyze the impact of different entropy estimation methods, we compare pruning results using these approaches on Llama3.1-8B and Mistral-7B-v0.3.

Figure~\ref{fig:entropy_estimation} presents the evaluation results across multiple benchmark datasets when deleting 12 layers of attention blocks using our method.
The results indicate that the choice of entropy estimation method significantly affects performance. Both Bucket-based and KNN-based estimation methods yield stable and high accuracy across all datasets, demonstrating their effectiveness in preserving essential model capabilities after pruning. In contrast, Renyi entropy estimation consistently underperforms, leading to noticeable accuracy degradation. This suggests that Renyi entropy may introduce excessive sensitivity to certain probability distributions, making it less suitable for pruning decisions of pre-trained Transformer blocks.

To investigate the redundancy in Transformer models, we analyze the impact of attention layer deletion across multiple datasets, including ARC-C, HellaSwag, MMLU, and WinoGrande. Figure~\ref{fig:layer_deletion} presents the performance degradation trend on MMLU as attention layers are progressively removed from Mistral-7B-v0.3. 
The results indicate that model performance remains stable until approximately 12 attention layers are removed, after which accuracy begins to degrade. This suggests that a significant portion of attention layers are redundant and can be pruned without substantial performance loss. Additionally, we compare different importance estimation methods for attention pruning. Both Bucket-based and KNN-based entropy estimation methods consistently outperform Cosine Similarity, demonstrating their effectiveness in identifying unimportant attention layers. In contrast, Renyi entropy performs poorly from the beginning, further confirming its limitations in guiding structured pruning.

\begin{figure}[t]
    \begin{center}
    \includegraphics[width=.9\linewidth]{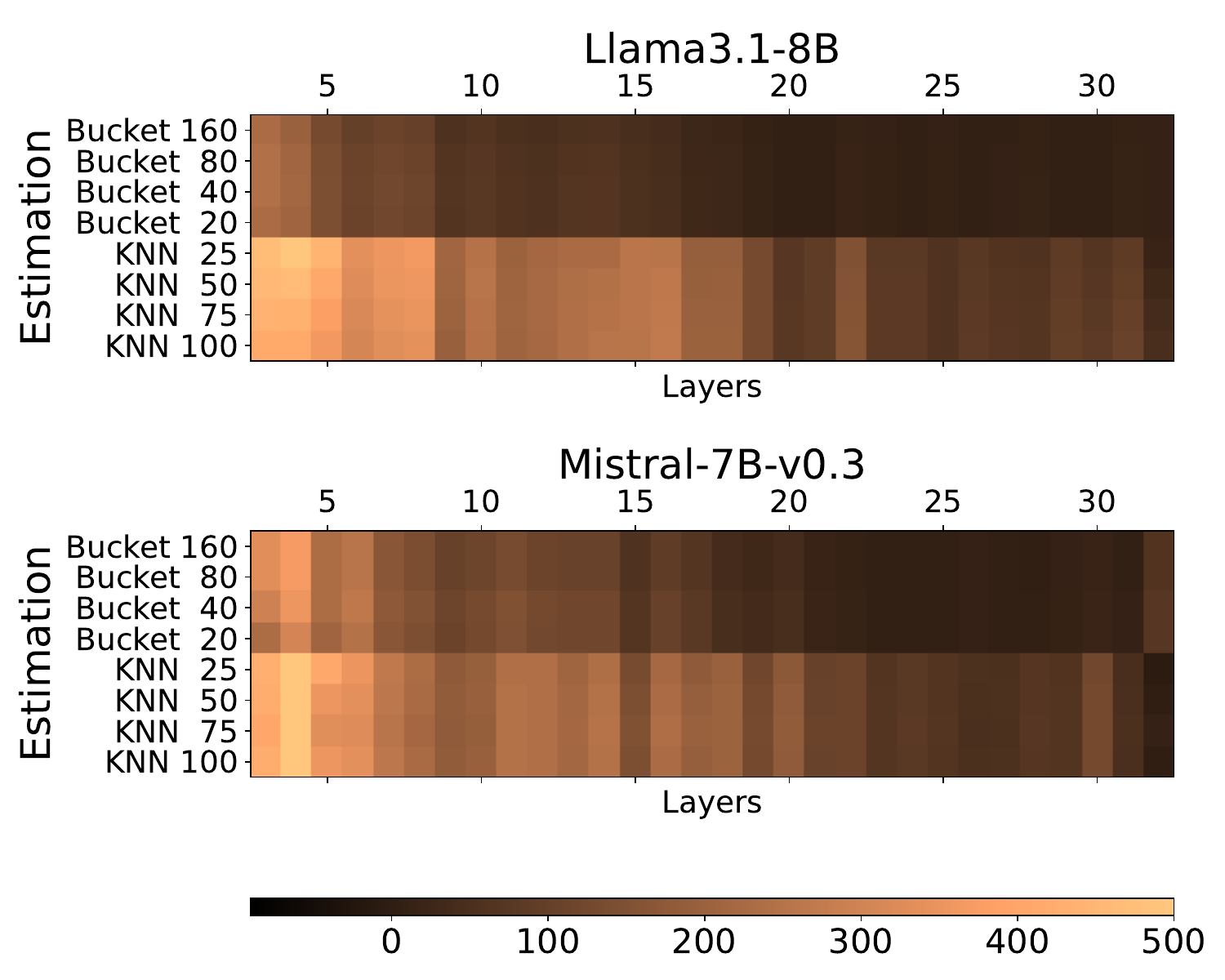}
    \end{center}
    \caption{Impact of Hyperparameter Variations}
    \label{fig:estimation_heatmap}
\end{figure}

\textbf{Estimation Hyper-parameter}. To further analyze the robustness of entropy estimation methods, we investigate the impact of different hyperparameter settings. Specifically, we tune the following parameters: Bucket-based Estimation: Number of bins selected from $\{20, 40, 80, 160\}$. KNN-based Estimation: Number of nearest neighbors selected from $\{25, 50, 75, 100\}$. Figure~\ref{fig:estimation_heatmap} presents the estimated entropy values across Transformer layers using different hyperparameter settings. The results indicate that while different entropy estimation methods (Bucket vs. KNN) yield significantly different absolute entropy values, the relative importance ranking of layers remains largely unchanged within same estimation method. This suggests that the choice of hyperparameter
(e.g., number of bins for Bucket-based estimation, number of neighbors for KNN-based estimation) does not significantly impact the identification of redundant layers.

The findings highlight the importance of selecting entropy estimation method while also reinforcing the stability of entropy-based pruning. Although different methods may compute varying absolute entropy values, the pruning decisions remain consistent for different methods. From our experiments, Bucket-based and KNN-based methods provide reliable performance, whereas using an inappropriate method like Renyi entropy could lead to suboptimal pruning outcomes.

\subsection{Speedup Test}
To evaluate the efficiency gains from pruning attention blocks, we conduct inference speed tests on Llama3.1-8B and Mistral-7B-v0.3. We prune attention layers progressively and measure both model performance and inference time. The speed test is performed by fixing the input sequence length to 1024 tokens and generating an output of 1024 tokens. Each experiment is repeated 10 times, and the average inference time is reported.

\begin{figure}[t]
    \begin{center}
    \includegraphics[width=0.85\linewidth]{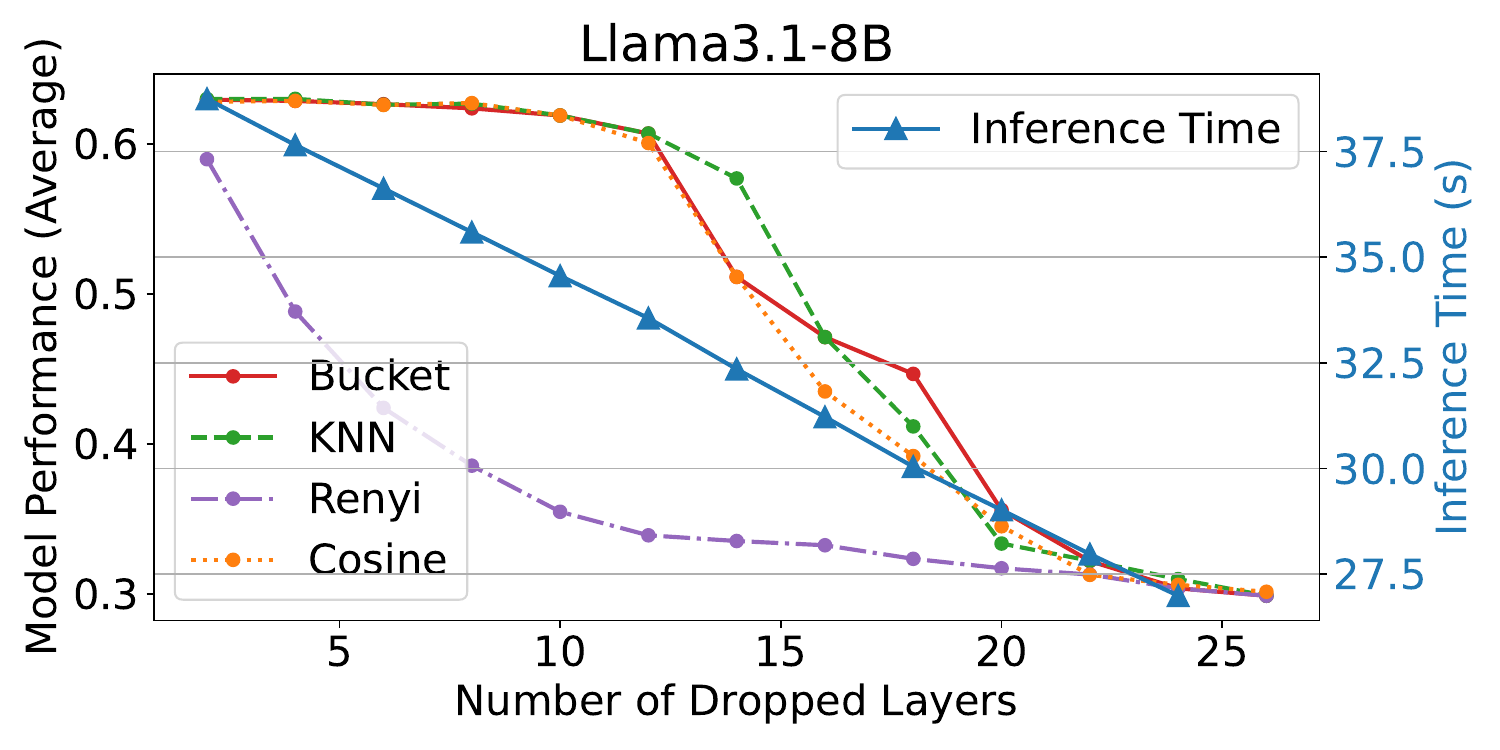}
    \includegraphics[width=0.85\linewidth]{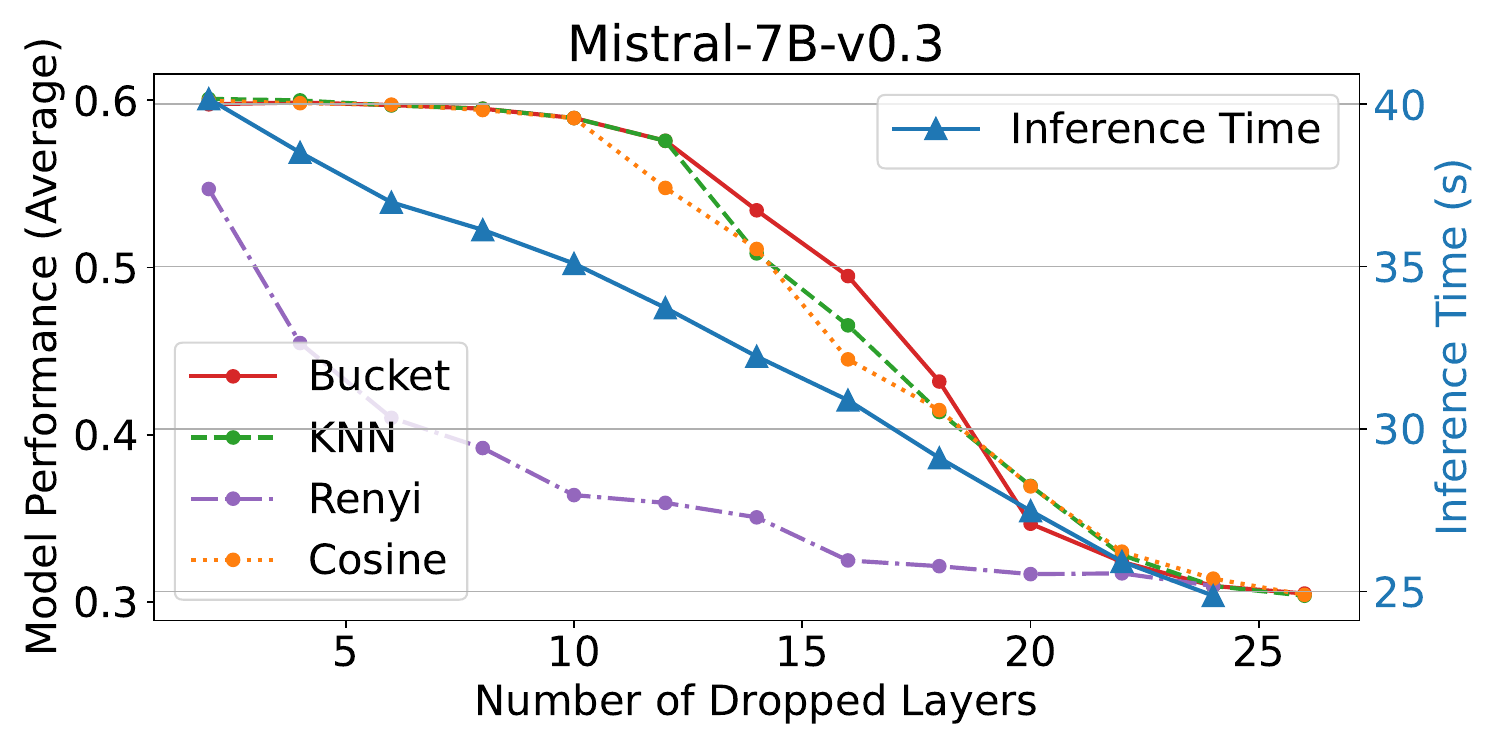}
    \end{center}
    \caption{SpeedUp Experiments}
    \label{fig:speedup}
\end{figure}

Figure~\ref{fig:speedup} shows the relationship between the number of dropped attention layers, model performance, and inference time. The results indicate that inference time decreases linearly as more attention layers are pruned. Notably, the first 12 layers provide the most significant speedup while maintaining model performance. Beyond this point, additional pruning begins to negatively impact accuracy. EntroDrop based on Bucket/KNN estimation outperforms currently widely used cosine similarity. Overall, these results highlight that our method can achieve substantial computational savings while preserving accuracy, making it an effective strategy for accelerating large language models in real-world deployment scenarios.

\section{Related Work}
\textbf{LLM Pruning.} In the era of large language models (LLMs), various methods have been proposed to reduce model size and accelerate inference~\citep{frantar2022gptq,lin2024awq,xiao2023smoothquant,shao2023omniquant,zhu2023survey,xu2023qa,dettmers2023spqr,liu2023llm}. Recent advances focus on post-training pruning techniques that eliminate redundant parameters or structures. SparseGPT~\citep{frantar2023sparsegpt} leverages second-order information to identify unimportant parameters in LLMs. Wanda~\citep{sun2023simple} introduces a pruning matrix that considers both weight magnitude and corresponding input activations. 
NEPENTHE~\cite{liao2024nepenthe} introduces a method that utilizes entropy to identify and remove low-entropy layers in deep neural networks, effectively reducing model depth while maintaining performance.
E-Sparse~\cite{li2023sparse} introduces an entropy-based pruning method that enhances inference speed and reduces memory usage in large language models by leveraging information richness to guide N:M sparsity. 
SPP~\citep{lu2024spp} designs an efficient fine-tuning method to recover model performance post-pruning while maintaining sparsity. Beyond parameter pruning, structural pruning of LLMs has also gained popularity. LLM-Pruner~\citep{ma2023llm} and ShearedLLaMA~\citep{xia2023sheared} remove unimportant structures such as layers and attention heads. Additionally, \cite{lu2024not} finds that certain experts in mixture-of-experts (MoE) LLMs can also be pruned. Among structural pruning methods, layer pruning is particularly relevant. Laco~\citep{yang-etal-2024-laco} reduces model depth by merging adjacent layers from the topmost layer downward. ShortGPT~\citep{men2024shortgpt} prunes unimportant layers based on a cosine similarity criterion. LLMDrop~\citep{he2024matters} finds that attention layers are more redundant than MLP layers but also relies on cosine similarity for pruning. Different from these approaches, in this paper, we propose a more effective criterion \textit{i.e.} \textbf{Entropy Increase} to identify and remove unimportant layers.

\section{Conclusion}
In this paper, we propose EntroDrop, an entropy-based pruning method that leverages entropy increase to identify and remove redundant computation blocks in large language models. Unlike traditional pruning approaches that rely on cosine similarity, our method captures the information flow within the model, leading to more effective pruning decisions. Through empirical analysis, we reveal distinct entropy dynamics across Transformer layers and demonstrate that entropy serves as a reliable metric for determining block importance.

\section{Limitations}
EntroDrop relies on a calibration dataset to estimate entropy dynamics. While we observe robustness across different datasets, its effectiveness in highly domain-specific tasks requires further exploration.
% Additionally, the impact of entropy-based pruning on fine-tuning and transfer learning requires further investigation. 
% Pruning may affect the model’s ability to generalize to new data distributions, potentially leading to performance degradation in downstream tasks. 
During the paper writing, generative AI tools (ChatGPT, Grammarly) are used to help fix grammatical issues and typos. As a paper tailored to efficient LLM deployment, we do not think any Potential Risks need to be addressed here. 

% \section*{Acknowledgments}

% \newpage
\bibliography{reference}

% \appendix

% \section{Example Appendix}
% \label{sec:appendix}

% This is an appendix.

\end{document}